\documentclass[journal]{IEEEtran}
\usepackage{latexsym,,amssymb,amsmath,graphicx,epsf,cite,bbm,float}
\usepackage{ifpdf}
\usepackage{epstopdf,mathtools}

\usepackage{algorithm,algorithmic}
\usepackage{amsmath,amssymb,bm}
\usepackage{amsfonts,dsfont,color,bbm,subcaption}

\def\ninept{\def\baselinestretch{1}}
\ninept

\newcommand{\abs}[1]{|#1|}

\DeclareMathOperator*{\argmin}{arg\,min}

\usepackage{hyperref,amsthm}

\newtheorem{theorem}{Theorem}

\newtheorem{lemma}[]{Lemma}

\newtheorem{proposition}[]{Proposition}

\newtheorem{corollary}[]{Corollary}

\newtheorem{remark}[]{Remark}

\newtheorem{definition}[]{Definition}

\begin{document}

\title{Nonparametric Extrema Analysis in Time Series for Envelope Extraction, Peak Detection and Clustering} 
\author{\IEEEauthorblockN{Kaan Gokcesu}, \IEEEauthorblockN{Hakan Gokcesu} }
\maketitle

\begin{abstract}
	In this paper, we propose a nonparametric approach that can be used in envelope extraction, peak-burst detection and clustering in time series. Our problem formalization results in a naturally defined splitting/forking of the time series. With a possibly hierarchical implementation, it can be used for various applications in machine learning, signal processing and mathematical finance. From an incoming input signal, our iterative procedure sequentially creates two signals (one upper bounding and one lower bounding signal) by minimizing the cumulative $L_1$ drift. We show that a solution can be efficiently calculated by use of a Viterbi-like path tracking algorithm together with an optimal elimination rule. We consider many interesting settings, where our algorithm has near-linear time complexities.
\end{abstract}

\section{Introduction}

\subsection{Extrema Analysis}
The problem of extrema analysis of time series \cite{hall2002moving,fink2011compression,vemulapalli2012robust}, whether in the form of envelope extraction \cite{tayfun1989wave}, peak-burst detection \cite{palshikar2009simple}, clustering \cite{kadambe1999wavelet} or signal bands/channels \cite{donchian1960commodities}
have become prominent in machine learning, data mining and mathematical finance. The identification and analysis of the extrema in a given time series is essential in many applications because of its useful topological implications and features \cite{palshikar2009simple} in a myriad of fields, including but not limited to mass spectrometry \cite{coombes2005improved}, signal processing \cite{jordanov2002digital,harmer2008peak}, image processing \cite{ma2005developing}, bioinformatics \cite{gokcesu2018adaptive,azzini2004simple,gokcesu2018semg} and astrophysics \cite{zhu2003efficient}.

The extrema can implicate certain increases (or decreases) in a time series, which can be used in the problem of classification \cite{gokcesu2021optimally}. For example, the workload (utilization) of a server CPU (processors) can be analyzed to extract idle or busy states, which can be used to forecast future states \cite{choi1996daily}.
The extrema analysis can efficiently track the nominal behavior of a time series, which can be used in the problem of anomaly detection \cite{gokcesu2018sequential,delibalta2016online,gokcesu2017online}. For example, the volumes of traffic in network data can be analyzed for anomalous or malicious behavior (such as attacks), where the analysis of periodicities or similarities can be meaningful \cite{vlachos2004identifying}.
The extrema analysis can model the dependencies in a time series, which can be used in the problem of online learning \cite{gokcesu2020generalized,neyshabouri2018asymptotically,gokcesu2020recursive}. For example, the trading volume and prices in financial data can be analyzed for the prediction of oversold or overbought markets, where their correlating burst events can be meaningful \cite{vlachos2008correlating}.
The extrema can track the most informative values in the time series, such as the envelope of the electromyography (EMG) signal for use in robotic systems and prosthesis control \cite{jang2016emg} to achieve a perfect collaboration between man and robot. 
In general, the extrema can provide useful robust features \cite{vemulapalli2012robust}, especially for robust learning \cite{gokcesu2021generalized}. 
The extrema are very important in these applications whether in the form of maxima or minima \cite{palshikar2009simple}. 

\subsection{Envelope Extraction}
The first of many applications of the extrema analysis is the envelope extraction, which is of utmost importance in the informative data analysis of a modulated signal, where the envelope is extracted by use of a demodulator \cite{bracewell1986fourier}. 

Moreover, its usefulness is most apparent in the analysis of random signals, which is commonly encountered in the study of seakeeping and oceanography (or ocean engineering), where waves often occur in groups, which can result in severe damage to the offshore structures. The theory of the envelope of a random signal have been historically used to investigate the statistical properties of such wave groups and correspondingly the wave amplitude distribution \cite{longuet1984statistical,ochi1989stochastic,tayfun1984nonlinear}. Their use is of high importance in the prediction of the deck motion (which is necessary for safe landing and take-off for helicopters), where the extraction of the deck motion envelope is used in the platform movement predictions \cite{tiecheng1991real}. The works of \cite{rice1944mathematical,rice1945mathematical} has established the basis for the envelope statistics and the expression for its spectra was obtained by \cite{tayfun1989wave}. The efficient calculation of the envelope is addressed in \cite{fang1995analysis}.

Furthermore, the envelope extraction problem has become prominent with the rise of the field of bioinformatics, especially the electrical activity of the human body and specifically the skeletal muscle tissue, i.e., the electromyography (EMG) signal. The most widely used techniques to extract the EMG envelope are the Hilbert transform \cite{chen2017feature}, mean square error (MSE) criterion \cite{d2001extraction} and the rectified signal waveform \cite{xie2006mean}. Many algorithms have been developed in literature to extract the EMG envelope, which primarily exploit the signal's moving average activity together with a noise reduction \cite{kleissen1994estimation}. However, such methods may suffer large bias errors and be unable to avoid measurement spikes \cite{marquez2019analysis}. The Savitsky-Golay smoother combined with a low-pass filter was used to address such problems \cite{yeh2016quantifying}, where the smoothing improves envelope shaping but introduces time delays, which may be intolerable in certain applications (such as robotics). In \cite{jang2016emg}, the signal envelope has been extracted with time-frequency analysis on finite time windows (using Short-time Fourier Transform). In \cite{myers2003rectification}, the authors suggest rectifying the EMG signal a priori, which can enhance the muscle activation information but still struggles with high envelope variability \cite{roberts2008interpreting}. In addition, Kalman and Wiener filter based algorithms have been developed for EMG by many works \cite{zhan2009filtering,lopez2009robust,tsui2009self,menegaldo2017real,triwiyanto2018muscle}. While the traditional Kalman setting has effective tracking when noise is white Gaussian; it provides no meaningful advantage against other methods, since EMG noise is strictly non-white \cite{marquez2019analysis}. Another approach by \cite{shmaliy2017unbiased} completely forgoes the unbiased noise assumption and is considered more robust \cite{shmaliy2018comparing}.

\subsection{Peak-Burst Detection}
Another application of interest for the extrema analysis is the well-studied peak (and burst) detection, which is the study of sudden increases or decreases in the data set for informative purposes. Note that while it is easy to visually identify peaks in small datasets or time-series, the notion of peak need to be formalized for objectivity and for the development of automated peak detection algorithms. The automation becomes important in applications involving large datasets and variables, such as the data center monitoring where there are many large time-series about the utilization of CPU and memory in many servers, which need to be analyzed in real-time. In \cite{palshikar2009simple}, several different notions of peak was introduced together with their corresponding algorithms.

Peak detection is a common task in many applications involving time series analysis. Standard approaches include either fitting a known function to the (possibly smoothed) time series; or matching a known peak shape in the time-series.
Another widely used approach to detect the peaks is the detection of the zero-crossings of the gradient in the time series (i.e., local extrema), which is found by analyzing the differences between a point and its neighbors for the change in the gradient sign. However, such methods are susceptible to errors in noisy observations, which needs to be solved by the use of a signal to noise ratio (SNR) threshold \cite{nijm2007comparison,jordanov2002digital}. The work in \cite{ma2005developing} parametrically learns such threshold by adapting it to the noise levels in the time-series.

It is also possible to analyze the time series according to different properties of a possible peak and detecting the ones which satisfy all these requirements. One such work is \cite{azzini2004simple}, where the analysis of peaks in gene expression microarray time series data (specifically, for malaria parasite Plasmodium falciparum) is studied using multiple methods, each of which assigns a score to every point of the time series. In one of the methods, the score is set as the rate of change (gradient) at each point. In another one, it is the fraction of the area under a certain candidate peak point. For each method, the top $10$ performing candidates are selected and the peaks detected by multiple methods are estimated as the true peaks, i.e., it is assumed that the true peaks need to satisfy more than one condition. In that application, the peaks detected are used to identify genes, and support vector machine (SVM) is used to assign each estimated gene to a group. The main problems encountered in peak detection are the existing noise in the time series and the fact that peaks can have varying amplitudes and scales, which may result in a high false positive rate \cite{palshikar2009simple}. 

There are some algorithms, which utilize the shape characteristics of the peaks in certain applications. One such approach is \cite{du2006improved}, which proposes a continuous wavelet transform (CWT) based algorithm to match preset pattern shapes to the possible peaks in mass spectroscopy data. The two dimensional CWT coefficients of the dataset is calculated using a Mexican-Hat mother wavelet (which has a basic peak shape) for the incoming time-series at multiple distinct scales. The changes in the wavelet transform are examined to identify possible peaks. Other wavelet based approaches to analyze spectroscopy data also exists \cite{coombes2005improved,lange2006high}.

Another approach is to analyze the rate of change in the time series. One example is \cite{harmer2008peak}, which proposes a 'momentum' based algorithm to detect possible peaks. The main idea is to compute the 'velocity', i.e., gradient, and 'momentum', i.e., the product of the functional value and the 'velocity' at various points. If a ball were to be dropped from a previous peak, it should gain momentum as it rolls down and similarly lose momentum as it climbs up the next peak. The point where it comes to rest and stop, i.e., when the momentum is zero, should be the next peak. Physical world analogs of the Newton's laws are proposed to calculate the changes in momentum as the ball follows along the time-series. 

The burst detection is also a prominent problem in machine learning, where the bursts are, in a sense, wider versions of the peaks. While a burst is made up of a wide region of extreme values with fast changes on both sides, a peak is made up of a narrow region of extreme values with fast changes on both sides. In \cite{zhu2003efficient}, the authors propose a wavelet based burst detection algorithm. The wavelet coefficients for Haar wavelets together with some time-domain statistics such as the windowed average are used in a data structure called the shifted wavelet tree (SWT). Each level of the tree represents a resolution (time scale) and each node of a level corresponds to a certain window. By iterating over different window sizes and time scales, the bursts are detected. The work in \cite{zhu2003efficient} apply this method to detect burst in Gamma ray time series in real-time for the Milagro astronomical telescope, where the burst durations vary greatly from minutes to days. 

The work in \cite{vlachos2004identifying} propose a moving average based algorithm for burst detection. The time-series is smoothed with a moving average filter and the values which are larger than some preset parameter $\theta$ times the standard deviation of the smoothed time-series are detected. They choose $\theta$ typically between $1.5$ and $2$. The smoothing window is selected by using domain knowledge (e.g., $30$ for daily data). In \cite{vlachos2008correlating}, the authors detect burst in real-time streaming stock market data and analyze correlations between the detected bursts. 

\subsection{Clustering}
The extrema analysis and specifically peak detection based approaches are also widely used in applications considering clustering (or unsupervised classification). One such popular application is the detection of the QRS complex in electrocardiogram (ECG) signals. Many different methods have been developed for detection of such events, which can be summarized as those based on wavelet transform \cite{coombes2005improved,du2006improved,fard2008novel,gregoire2011wavelet,kadambe1999wavelet,
	nenadic2004spike,singh2011robust
}, traditional window-threshold techniques \cite{pan1985real,jacobson2001auto,
excoffier1982automatic}, Hilbert transform \cite{benitez2001use}, combining Hilbert and wavelet transform \cite{rabbani2011r}, artificial neural networks \cite{xue1992neural,vijaya1998ann}, techniques using templates \cite{mtetwa2006smoothing,andreev2003universal}, morphology filtering \cite{zhang2009qrs,zhang2011qrs}, nonlinear filtering \cite{sun1992microcontroller,ferdi2003r,aboy2005automatic,shim2009nonlinear}, Kalman filtering \cite{tzallas2006epileptic}, Gabor filtering \cite{nguyen2009peak}, Gaussian second derivative filtering \cite{fredriksson2009automatic}, linear prediction analysis \cite{lin1989qrs}, higher-order statistics \cite{panoulas2001enhancement}, K-Means clustering \cite{mehta2010k}, fuzzy C-Means clustering \cite{sharma2011development}, Empirical Mode Decomposition \cite{slimane2010qrs}, hidden Markov models \cite{coast1990approach}, and techniques using histogram/cumulative distribution function \cite{sezan1990peak}, intensity weighted variance \cite{jarman2003new}, stochastic resonance \cite{deng2006linear}, or a smoothed nonlinear energy operator \cite{mukhopadhyay1998new}.

\subsection{Signal Bands}
The time-series bands or channels is another (maybe most widely used) application (especially in the stock market). Within economics and mathematical finance, the trading bands are used for a variety of different trading approaches because of the various studies in support of them
\cite{treynor1985defense,brock1992simple,cowie2011evaluation}.

A trading band is, in its most simple form, an envelope (or channel) around a certain evolving metric (property) of the time series. Traditionally, these kinds of bands are constructed by creating two signals that are individually a certain distance away from a centralizing signal, which consequently define an upper and a lower band. The distance between the upper and lower bands is typically dependent on the trading band type, which is usually either a measure of volatility or a function of some other related parameter \cite{john2002bollinger}. The upper and lower bands could be considered as some forms of support and resistance lines; where support and resistance are the values from which a quantity struggles to descend and ascend respectively \cite{murphy1999study}.

One of the most popular signal bands/channels is the Donchian Price Channels \cite{donchian1960commodities}, which defines a simple band with a set of three bands in total: one upper, one lower and a middle band, which is the direct average of them. For a preset window size $n$, the upper and the lower bands are the maximum and the minimum respectively of the last $n$ samples.
Bollinger bands are another popular technical analysis tool in mathematical finance. Aside from the field of economics, Bollinger bands have also been used to measure the accident rate (as a safety indicator) in the air travel industry and as a method of inspection in patterned fabrics \cite{cowie2011evaluation}. Bollinger bands consist of two main values, which are a measure of the central tendency together with a measure of the volatility \cite{john2002bollinger}. The traditionally used central tendency measure is the moving average $\mu_t$ and volatility measure is the moving standard deviation $\sigma_t$. These two values create three bands in total, where the middle band is the moving average $\mu_t$ and the upper and lower bands are $\mu_t\pm 2\sigma_t$.
Another popular signal band is Keltner channels, which again uses the moving average for the measure of the central tendency. However, for the measure of the volatility, they adopt the Average True Range (ATR) \cite{wilder1978new}. 

\subsection{Contributions and Organization}
The problem with most of the algorithms in literature is in their parametric nature, where the more generally applicable and nonrestrictive they are, the more free parameters they have \cite{scholkmann2012efficient,palshikar2009simple}. However, the selection of parameters not only depend on the application and the dataset, it is also non trivial. To this end, we propose a nonparametric approach for the extrema analysis problem, which can be used in many types of applications including envelope extraction, peak-burst detection and clustering. In \autoref{sec:formulation}, we mathematically formalize the problem setting in a natural way. In \autoref{sec:method}, we provide the methodology to find the solution for that formulation, which sequentially outputs two non-crossing time series for a given input time series. In \autoref{sec:analysis}, we provide the computational complexity analysis for many interesting settings and show its efficiency. In \autoref{sec:conclusion}, we provide some concluding remarks.  

\section{Problem Formulation}\label{sec:formulation}
In this section, we formalize the extrema analysis problem for a given signal, mathematically. We start with an incoming time series, i.e., $\{x_1,x_2,\ldots,x_T\}$, which is written for brevity as $\{x_t\}_{t=1}^T$. For this incoming time series, we want to create two signals, which are $\{a_t\}_{t=1}^T$ and $\{b_t\}_{t=1}^T$. These two signals will be our upper and lower bounding time series respectively. Thus, we want $a_t\geq x_t$ and $b_t\leq x_t$ for all $t$.

We postulate that there is a natural nonparametric way of defining the upper and lower bounding time series. For a definition that is parameter-free, we create the sequences $\{a_t\}_{t=1}^T$ and $\{b_t\}_{t=1}^T$ directly from the time series $x_t$, such that at any time $t$, $x_t$ is either $a_t$ or $b_t$. Hence, the problem is to split the time series $\{x_t\}_{t=1}^T$ into two sequences $\{a_t\}_{t=1}^T$ and $\{b_t\}_{t=1}^T$ by finding a binary sequence $\{s_t\}_{t=1}^T$, for which
\begin{align}
	a_t=&s_tx_t+(1-s_t)a_{t-1},\\
	b_t=&(1-s_t)x_t+s_tb_{t-1}.
\end{align}
The sequence $\{s_t\}_{t=1}^T$ is selected by minimizing the following objective:
\begin{align}
	\min_{\{s_t\}_{t=1}^T}\left(\sum_{t=2}^{T}\abs{a_t-a_{t-1}}+\sum_{t=2}^T\abs{b_t-b_{t-1}}\right).\label{eq:loss}
\end{align}
Hence, the problem is to split $x_t$ into two sequences such that the sum of their $L_1$ (Manhattan) drifts (sum of the successive samples distances) are minimized.

We have constructed the envelopes $\{a_t\}_{t=1}^T$ and $\{b_t\}_{t=1}^T$ with piecewise constant interpolation for the formulation, where $a_t$ (and similarly $b_t$) is either $x_t$ or $x_{\tau}$ for some $\tau<t$. However, we can just as easily use linear interpolation (which is more common and intuitive) for the missing the indices, which does not change the $L_1$ drift (and the objective function). With linear interpolation, we have the following beautiful property.

\begin{lemma}
	The signals $\{a_t\}_{t=1}^T$ and $\{b_t\}_{t=1}^T$ never cross.
	\begin{proof}
		Suppose they cross at some $t_c$, which is a real number and both $a$ and $b$ are interpolated there. For that, let us have $t_{a,1}<t_c<t_{a,2}$ and $t_{b,1}<t_c<t_{b,2}$, where $a_{t_{a,1}}=x_{t_{a,1}}$, $a_{t_{a,2}}=x_{t_{a,2}}$, $b_{t_{b,1}}=x_{t_{b,1}}$, $b_{t_{b,2}}=x_{t_{b,2}}$ and $a_t$ and $b_t$ are interpolated between the samples $\{t_{a,1},t_{a,2}\}$ and $\{t_{b,1},t_{b,2}\}$ respectively. This will have the following loss
		\begin{align}
		L_T=&\sum_{t=2}^{t_{a,1}}\abs{a_t-a_{t-1}}+\sum_{t=t_{a,2}}^{T}\abs{a_t-a_{t-1}}+\abs{a_{t_{a,1}}-a_{t_{a,2}}},\nonumber\\
		&+\sum_{t=2}^{t_{b,1}}\abs{b_t-b_{t-1}}+\sum_{t=t_{b,2}}^{T}\abs{b_t-b_{t-1}}+\abs{b_{t_{b,1}}-b_{t_{b,2}}}.
		\end{align}
		Now, if we switch the classes of the samples after $t_c$, i.e., $a_{t\geq t_{b,2}}\leftarrow b_{t\geq t_{b,2}}$ and vice-versa, we will have a loss which has a $\triangle$ difference resulting from the cross at $t_c$, which is $\triangle=\abs{a_{t_{a,1}}-b_{t_{b,2}}}+\abs{b_{t_{b,1}}-a_{t_{a,2}}}-\abs{a_{t_{a,1}}-a_{t_{a,2}}}-\abs{b_{t_{b,1}}-b_{t_{b,2}}}$.
		Since the original $a_t$ and $b_t$ were crossing each other, we want to compare the sum of the diagonals of a butterfly with the sum of its sides. From the triangle inequality, we have $\triangle<0$, i.e., a smaller loss, which concludes the proof.
	\end{proof}
\end{lemma}

\section{Methodology}\label{sec:method}
\subsection{Path Tracking Algorithm}
The brute force approach is to try for all possible $\{s_t\}_{t=1}^T$ sequence, which are $2^T$ in total (i.e., exponential), and thus intractable.

A tractable approach is to use a path tracking method similar in spirit to the Viterbi algorithm \cite{viterbi1967error}. In this approach, we sequentially calculate the losses of the sequences (or paths) $\{s_t\}_{t=1}^T$ and only keep the paths with the minimal losses (the survival rule). 

\begin{definition}\label{thm:twos_t}
	At any time $t'$, let $\{s_{1,t}\}_{t=1}^{t'}$ and $\{s_{2,t}\}_{t=1}^{t'}$ be two distinct sequences, which are identical after some time $t_s$, such that
	\begin{itemize}
		\item $s_{1,t}=s_{2,t}=b$ for $t_s<t\leq t'$ (where $b\in\{0,1\}$), 
		\item $s_{1,t_s}=s_{2,t_s}=1-b$,
		\item $\{s_{1,t}\}_{t=1}^{t_s-1}$ is distinct to $\{s_{2,t}\}_{t=1}^{t_s-1}$.
	\end{itemize} 
\end{definition}

\begin{proposition}
	For two sequences $\{s_{1,t}\}_{t=1}^{t'}$ and $\{s_{2,t}\}_{t=1}^{t'}$ as in \autoref{thm:twos_t}, let their losses be $L_{1,t'}$ and $L_{2,t'}$. We need only to keep track of the sequence with the smaller loss. 
	\begin{proof}
		We observe that the last time $\{s_{1,t}\}_{t=1}^{t'}$ and $\{s_{2,t}\}_{t=1}^{t'}$ is $0$ and $1$ coincide, thus, the additive loss resulting from the new sample $x_{t'+1}$ is identical. Without loss of generality, assume that the loss of $\{s_{1,t}\}_{t=1}^{t'}$ up to $t'$, $L_{1,t'}$ is smaller. Then, there is no reason to keep track of the loss of $\{s_{2,t}\}_{t=1}^{t'}$, since for any sequence $\{s_t\}_{t=t'+1}^T$, the loss of $\{s_{1,t}\}_{t=1}^{t'}\cup\{s_t\}_{t=t'+1}^T$ will be smaller than the loss of $\{s_{2,t}\}_{t=1}^{t'}\cup\{s_t\}_{t=t'+1}^T$, which concludes the proof. 
	\end{proof}
\end{proposition}

Using this property, we keep track of the sequences with equivalence classes. At any time $t'$, $A_{t'};b_{\tau}$ for $\tau<t'$ denote the class of the sequences $\{s_{t}\}_{t=1}^{t'}$, where the last index of $1$ is $t'$ and the last index of $0$ is $\tau$. Similarly, we have $B_{t'};a_{\tau}$.
At any time $t$, we will have $2(t-1)$ classes with the corresponding losses, which are given as
\begin{align}
	&A_t; b_{\tau<t} : L^{a}_{\tau<t},\\	
	&B_t; a_{\tau<t} : L^{b}_{\tau<t}.	
\end{align}
Note that each class keeps track of a single surviving sequence with the best loss. With a new incoming sample $x_{t+1}$, we update the classes and their losses as follows:
\begin{align}
	&A_{t+1}; b_{\tau<t} &&: L^a_{\tau<t}+\abs{x_{t+1}-x_t},\\
	&A_{t+1}; b_t &&: \min_{\tau<t}(L^b_{\tau}+\abs{x_{t+1}-x_\tau}),\label{At+1bt}\\
	&B_{t+1}; a_{\tau<t} &&: L^b_{\tau<t}+\abs{x_{t+1}-x_t},\\
	&B_{t+1}; a_t &&: \min_{\tau<t}(L^a_{\tau}+\abs{x_{t+1}-x_\tau}),\label{Bt+1at}
\end{align}
The minimum operations at \eqref{At+1bt} and \eqref{Bt+1at} are where the Viterbi behavior comes from.
At the end of $T$ rounds, we will have
\begin{align}
&A_{T}; b_{\tau<T} &&: L^a_{\tau<T},\\
&B_{T}; a_{\tau<T} &&: L^b_{\tau<T},
\end{align}
We can select the class with the minimum cumulative loss, which will minimize \eqref{eq:loss}. This approach has quadratic $O(T^2)$ computational and memory complexity.

\subsection{Efficient Algorithm}
First of all, we observe that keeping both the equivalence classes $A_t;b_{\tau}$ and $B_t;a_{\tau}$ at time $t$ is unnecessary (since there is no distinction between them). To this end, we revamp the design and only keep the following losses at each time $t$:
\begin{align}
	L_{t,\tau}, && t\in{0,1,\ldots,t-1},
\end{align}
where $L_{t,\tau}$ represents the loss of the best sequence where $\{s_{t'}\}_{t'=\tau+1}^t$ is $b\in\{0,1\}$ and $s_\tau=1-b$, i.e., if $\{x_{t'}\}_{t'=\tau+1}^t$ belongs to the time series $a$ then $x_\tau$ belongs to the time series $b$ and vice-versa (thus, at $\tau$, the assignment changes).
The sequential update of $L_{t,\tau}$ is given as the following:
\begin{align}
	L_{t+1,\tau}&=L_{t,\tau}+\abs{x_{t+1}-x_t}, \tau\in\{0,1,\ldots,t-1\},\\
	L_{t+1,t}&=\min(L_{t,0},\min_{1\leq\tau<t}(L_{t,\tau}+\abs{x_{t+1}-x_\tau})).
\end{align}
We observe that the individual updates to $L_{t+1,\tau}$ for $\tau\in\{0,1,\ldots,t-1\}$ is unnecessary since we only need to compare the losses at the end. Thus, we modify it as the following:
\begin{align}
M_{t+1,\tau}=\begin{cases}
M_{t,\tau}, &\tau\in\{0,1,\ldots,t-1\},\\
\displaystyle\min_{0\leq\tau'<t}(M_{t,\tau'}+d_{t,\tau'}), &\tau=t
\end{cases}
\end{align}
where $d_{t,0}=-\abs{x_{t+1}-x_t}$, $d_{t,\tau}=-\abs{x_{t+1}-x_t}+\abs{x_{t+1}-x_\tau}$.

Note that we have $M_{1,0}=0$, $M_{t,0}=0$.
For further efficiency, we also keep the minimizer $\tau$ at each time $t$
\begin{align}
	\tau_t=\argmin_{0\leq\tau<t}(M_{t,\tau}+d_{t,\tau}).
\end{align}
At the end, we recreate the sequence as follows:
We calculate
\begin{align}
	\tau^*=\argmin_{0\leq\tau\leq T-1} M_{T,\tau},
\end{align}
which means the samples between $t\in\{\tau^*+1,\ldots,T\}$ belong to the first time series. Then, we get the next index, which is $\tau_{\tau^*}$, which means the samples between $t\in\{\tau_{\tau^*}+1,\ldots,\tau^*\}$ belong to the second time series. Then, setting $\tau^*\leftarrow\tau_{\tau^*}$, we recursively recalculate the time indices and setting the samples in the corresponding time series alternatively.

We can see that the double indexing is unnecessary since
\begin{align}
M_{t,t'}=M_{t'+1,t'}=\min_{0\leq\tau<t'}(M_{t',\tau}+d_{t',\tau}),	
\end{align}
Thus, we can remove one of the indexing and just write that
\begin{align}
M_t=\min_{0\leq\tau\leq t-1}(M_{\tau}+d_{t,\tau}),&&\tau_t=\argmin_{0\leq\tau\leq t-1}(M_{\tau}+d_{t,\tau})\label{Mt,taut}
\end{align}
where
\begin{align}
d_{t,\tau}=\begin{cases}
-\abs{x_{t+1}-x_t}+\abs{x_{t+1}-x_\tau}, &1\leq\tau\leq t-1\\
-\abs{x_{t+1}-x_t}, &\tau=0
\end{cases}.
\end{align}

This approach has $O(T)$ memory complexity and $O(T)$ backtrack. However, it will still have $O(T^2)$ complexity in total because of the calculation in $M_{t}$. 

\subsection{Optimal Elimination}
We observe that the reason for the quadratic time complexity is because the complexity at each time $t$ is proportional with the number of equivalence classes at time $t$. To make it efficient, we can eliminate certain equivalence classes. Sub-optimal ways include the elimination of $M_{\tau}$ where $\tau\leq t-w$ for some window $w$, or similarly, keeping the best $w$ classes (with the minimum $M_{\tau}$).  

We observe that there is a trimming approach to the equivalence classes without loss of optimality.

\begin{lemma}\label{thm:elimination}
	We observe that we can directly eliminate all $M_{\tau_1}$ for $\tau_1\neq 0$, such that
	\begin{align*}
	M_{\tau_1}\geq M_{\tau_2}+\abs{x_{\tau_2}-x_{\tau_1}},
	\end{align*}
	for some $\tau_2\notin\{0,\tau_1\}$. 
	\begin{proof}
		From the algorithm, we know that for any time $t$, if
		\begin{align}
		t^*=\argmin_{0\leq\tau\leq t-1}(M_{\tau}+d_{t,\tau}),
		\end{align}
		we have 
		\begin{align}
		M_t=M_{t^*}+d_{t,t^*}.
		\end{align}
		Let $\tau_1,\tau_2<t$. Then, we have
		\begin{align}
			M_{\tau_1}\geq& M_{\tau_2}+\abs{x_{\tau_2}-x_{\tau_1}},\\
			\geq&M_{\tau_2}+\abs{x_{t}-x_{\tau_2}}-\abs{x_{t}-x_{\tau_1}},
		\end{align}
		from the triangle inequality. Thus,
		\begin{align}
			M_{\tau_1}+d_{t,\tau_1}\geq M_{\tau_2}+d_{t,\tau_2}.
		\end{align}
		Thus there exists a $t^*$, which is distinct from $\tau_1$. Hence, we can eliminate $M_{\tau_1}$, which concludes the proof.
	\end{proof}
\end{lemma}

\begin{theorem}\label{thm:survival}
	If, for some $t_1<t_2$, 
	\begin{align*}
		x_{t_1}\in[\min(x_{t_2},x_{t_2+1}),\max(x_{t_2},x_{t_2+1})],
	\end{align*}
	$M_{t_1}$ does not survive.
	\begin{proof}
		From \autoref{thm:elimination}, for survivability, we require
		\begin{align}
		M_{t_1}<M_{t_2}+\abs{x_{t_1}-x_{t_2}},\label{condition}
		\end{align}
		for all nonzero $t_1\neq t_2$ pairs in the running.
		If $t_1<t_2$, we have 
		\begin{align}
		M_{t_2}\leq&M_{t_1}-\abs{x_{t_2+1}-x_{t_2}}+\abs{x_{t_2+1}-x_{t_1}},\label{Mt2Mt1}
		\end{align}
		from \eqref{Mt,taut}; and
		\begin{align}
			M_{t_1}-\abs{x_{t_1}-x_{t_2}}\leq M_{t_1}-\abs{x_{t_2+1}-x_{t_2}}+\abs{x_{t_2+1}-x_{t_1}}\label{Mt1tri}
		\end{align}
		from triangle inequality. However, \eqref{Mt1tri} holds with equality, i.e.,
		\begin{align}
			M_{t_1}-\abs{x_{t_1}-x_{t_2}}= M_{t_1}-\abs{x_{t_2+1}-x_{t_2}}+\abs{x_{t_2+1}-x_{t_1}}
		\end{align}
		when $x_{t_1}$ is between $x_{t_2}$ and $x_{t_2+1}$, which, together with \eqref{Mt2Mt1} results in
		\begin{align}
			M_{t_1}-\abs{x_{t_1}-x_{t_2}}\geq M_{t_2}.
		\end{align}
		Thus, $M_{t_{1}}$ breaks the condition in \eqref{condition}, which concludes the proof.
	\end{proof}
\end{theorem}

\section{Complexity Analysis}\label{sec:analysis}
In this section, we analyze the computational complexity of our optimal trimming. We analyze the number of surviving equivalence classes since the computational complexity per time is linearly dependent on that. 
While the complexity is straightforward to see for periodic signals, random processes need more analysis.
\subsection{Uniform i.i.d. Process}
Let $x_t$ be a uniform i.i.d. process such that $x_t\sim U([0,1])$, i.e., $f(x)=1$ for $x\in[0,1]$ and $0$ otherwise.

\begin{corollary}
	For a uniform i.i.d. process, the expected computational complexity of our algorithm is $O(\log T)$ per time.
	\begin{proof}
		At any arbitrary time $T$, let $\tau=T-t$ be the number of samples that come after $x_t$. From \autoref{thm:survival}, the only way $M_t$ will survive at time $T$ is if all the samples $x_{t'>t}$ are in $[0,x_t)$ or in $(x_t,1]$. Thus, the probability that $M_t$ survives is
		\begin{align}
		P_t=x_t^\tau+(1-x_t)^\tau,
		\end{align}
		where $\tau=T-t$. Since the survival is a Bernoulli trial with $P_t$, the expected survivability $z_t$ of $M_t$ is
		\begin{align}
		z_t\triangleq&\mathbb{E}_{x_t}[P_t],\\
		=&\int_{0}^{1}\left(x_t^\tau+(1-x_t)^\tau\right) dx_t,\\
		=& \left(\frac{x_t^{\tau+1}}{\tau+1}-\frac{(1-x_t)^{\tau+1}}{\tau+1}\right)\bigg\rvert_0^1,\\
		=&\frac{2}{\tau+1}.
		\end{align}
		Hence, the expected number of survivors $S_T$ at time $T$ is
		\begin{align}
		S_T=O(\log T),
		\end{align}
		which concludes the proof.
	\end{proof}
\end{corollary}

\subsection{Arbitrary i.i.d. Process}
 Let $x_t\in\Re$ be i.i.d. with $x_t\sim \mathcal{D}$ for some distribution $\mathcal{D}$. Let its probability density function (PDF) be $f(x)$ and its cumulative distribution function (CDF) be $F(x)$. 
 \begin{corollary}
 	For an arbitrary i.i.d. process, the expected complexity of our algorithm is $O(\log T)$ per time.
 	\begin{proof}
 		For $M_t$ to survive, we again need all of the upcoming samples $x_{t'>t}$ be strictly less or more than $x_t$. Thus, the probability of survival is
 		\begin{align}
 		P_t= (F(x_t))^\tau+(1-F(x_t))^\tau,
 		\end{align}  
 		and its expectation is
 		\begin{align}
 		z_t=&\int_\Re f(x_t) P_tdx_t,\\
 		=&\int_\Re f(x_t)\left[\left(F(x_t)\right)^\tau+\left(1-F(x_t)\right)^\tau\right]dx_t,\\
 		=&\int_0^1 \left[\left(F\right)^\tau+\left(1-F\right)^\tau\right]dF,\\
 		=&\frac{2}{t+1}.
 		\end{align}
 		Hence, the expected number of survivals at $T$ is $O(\log T)$.
 	\end{proof}
 \end{corollary}

\subsection{Simple Symmetric Random Walk}
Let $x_t$ be a simple symmetric random walk, where the successive difference 
\begin{align}
	\delta_{t}\triangleq x_{t}-x_{t-1},
\end{align}
is an unbiased random variable in $\{-1,1\}$, i.e.,
\begin{align}
	\delta_{t}=\begin{cases}
	+1, &\text{with probability $0.5$}\\
	-1, &\text{with probability $0.5$}
	\end{cases}.
\end{align}

\begin{corollary}
	For a simple symmetric lattice random walk with step size $1$, we have $O(\sqrt{T})$ complexity.
	\begin{proof}
		At time $T$, the survivability given $x_T$, $z_t$, of $M_t$ is again the sum of the probabilities of $x_t$ being greater or less than all of the following samples, which is given by
		\begin{align}
			z_t=\begin{cases}
			\mathbb{P}(x_{\tau}>x_t; t<\tau\leq T ), & x_T>x_t\\
			\mathbb{P}(x_{\tau}<x_t; t<\tau\leq T), &x_T<x_t
			\end{cases},
		\end{align}
		where $\mathbb{P}(\cdot)$ is the probability operator. From the Ballot theorem \cite{renault2007four}, we have
		\begin{align}
		z_t=\frac{\abs{x_T-x_t}}{T-t}.
		\end{align}
		Since the sum $\sum_{t'=t+1}^{T}\delta_{t'}$ converges in distribution to a Gaussian with variance $T-t$, the expectation of its absolute is $O(\sqrt{T-t})$. Thus, the total number of expected survivors $S_T$ at $T$ is
		\begin{align}
		S_T=O(\sqrt{T}),
		\end{align}
		which concludes the proof.
	\end{proof}
\end{corollary}
\begin{remark}
	The complexity bound will be the same for any arbitrary step size $\delta$ since we only compare whether the values $x_t$ are greater (or less) than the following samples $x_{\tau>t}$.
\end{remark}

\subsection{General Symmetric Random Walk}
Let $x_t$ be a random walk, where the successive difference 
\begin{align}
\delta_{t}\triangleq x_{t}-x_{t-1},
\end{align}
is an unbiased, finite variance random variable with a distribution $\mathcal{D}$, i.e.,
\begin{align}
\delta_t\sim D, &&\mathbb{E}[\delta_t]=0, &&\mathbb{E}[\delta_t^2]<\infty.
\end{align}
\begin{corollary}
	For a symmetric random walk with an unbiased, finite variance random step size $\delta$, we have $O(\sqrt{T})$ complexity.
	\begin{proof}
		Similarly to the simple random walk, the survivability $z_t$ of $M_t$ is
		\begin{align}
			\mathbb{E}[s_t]=O\left(\frac{1}{\sqrt{T-t}}\right),
		\end{align}
		from the extended Ballot theorem \cite{addario2008ballot}. Thus, the expected number of survivors at time $T$ is
		\begin{align}
			S_T=O(\sqrt{T}),
		\end{align}
		which concludes the proof.
	\end{proof}
\end{corollary}

\subsection{All-time Maximum and Minimum}
Another interesting setting is, let $x_t$ be the all time maximum with probability $p_t$ and all time minimum with probability $q_t$, i.e.,
\begin{align}
	\mathbb{P}(x_t\geq x_\tau; \tau<t)=&p_t,\\
	\mathbb{P}(x_t\leq x_\tau; \tau<t)=&q_t,
\end{align} 
where $\mathbb{P}(\cdot)$ is again the probability operator.
\begin{corollary}
	If at any time $t$, $x_t$ is the all time maximum with probability $p_t$ and all time minimum with probability $q_t$, where $p_t$ and $q_t$ are nonincreasing, i.e., $p_t\leq p_{t-1}$, $q_t\leq q_{t-1}$; the expected number of survivors is
	\begin{align}
		S_T=O\left(p_T^{-1}+q_T^{-1}\right).
	\end{align}
	
	\begin{proof}
		At time $T$, let $t_1$ be the last time we get an all time maximum and $t_2$ be the last time we get an all time minimum. Then, the number of surviving sequences are bounded by
		\begin{align}
		\sum_{t=1}^{T}z_t\leq& \max(t'_{1,T},t'_{2,T})\leq t'_{1,T}+t'_{2,T},
		\end{align}
		where $t'_{1,T}\triangleq T-t_1$ and $t'_{2,T}\triangleq T-t_2$. 
		Since the all time maximum and minimum probabilities are $p_t$ and $q_t$. Then, at time $T$, the expectation of $t'_{1,T}$ is
		\begin{align}
		\mathbb{E}[t'_{1,T}]=&p_T
		\begin{multlined}[t]
		+2(1-p_T)p_{T-1}\\
		+3(1-p_T)(1-p_{T-1})p_{T-2}\ldots,
		\end{multlined}\\
		\leq&1+(1-p_T)+(1-p_T)(1-p_{T-1})+\ldots,
		\end{align}
		where we have the following recursion
		\begin{align}
		\mathbb{E}[t'_{1,T}]\leq 1+(1-p_T)\mathbb{E}[t'_{1,T-1}],
		\end{align}
		and a similar recursion for $t'_{2,T}$ as well. Thus,
		\begin{align}
			\mathbb{E}[t'_{1,T}]\leq \sum_{i=0}^{T}(1-p_T)^i\leq \frac{1}{p_T},
		\end{align}
		for nonincreasing $p_t$ (similarly for $\mathbb{E}[t'_{2,T}]$ and $q_T$), which concludes the proof.
	\end{proof}
\end{corollary}

\section{Conclusion}\label{sec:conclusion}
	In this paper, we proposed a nonparametric approach that can be used in envelope extraction, peak-burst detection and clustering in time series. Our proposed problem formulation naturally defines a splitting/forking of the time series, which can be used for various applications in machine learning, signal processing and mathematically finance (with a possibly hierarchical implementation via successive application of our method to the generated forks). We proposed an algorithm that sequentially creates two signals (one upper bounding and one lower bounding signal) from an incoming input signal by minimizing the cumulative $L_1$ drift. We showed that a solution can be efficiently calculated by use of a Viterbi-like path tracking algorithm and in many interesting settings, our algorithm has low time complexities.

\bibliographystyle{IEEEtran}
\bibliography{double_bib}

\end{document}